\documentclass[10pt,twocolumn,letterpaper]{article}

\usepackage[pagenumbers]{cvpr} 

\usepackage{graphicx}
\usepackage{amsmath}
\usepackage{amssymb}
\usepackage{booktabs}
\usepackage{caption,subcaption}
\usepackage[linesnumbered, ruled, vlined]{algorithm2e}

\SetCommentSty{mycommfont}

\makeatletter
\patchcmd{\@algocf@start}
{-1.5em}
{0pt}
{}{}
\makeatother

\graphicspath{{imgs/}}

\usepackage[pagebackref,breaklinks,colorlinks]{hyperref}

\usepackage[capitalize]{cleveref}
\crefname{section}{Sec.}{Secs.}
\Crefname{section}{Section}{Sections}
\Crefname{table}{Table}{Tables}
\crefname{table}{Tab.}{Tabs.}

\def\cvprPaperID{4679} 
\def\confName{CVPR}
\def\confYear{2022}

\begin{document}
	\newcommand{\comm}[1]{}
	
	\title{Neural Architecture Search using Progressive Evolution}
	
	\author{Nilotpal Sinha\\
		National Yang Ming Chiao Tung University\\
		Hsinchu City, 30010, Taiwan\\
		{\tt\small nilotpalsinha.cs06g@nctu.edu.tw}
		\and
		Kuan-Wen~Chen\\
		National Yang Ming Chiao Tung University\\
		Hsinchu City, 30010, Taiwan\\
		{\tt\small kuanwen@cs.nctu.edu.tw}
	}
	\maketitle
	
	\begin{abstract}
		Vanilla neural architecture search using evolutionary algorithms (EA) involves
		evaluating each architecture by training it from scratch, which is extremely
		time-consuming. This can be reduced by using a supernet to estimate the fitness
		of every architecture in the search space due to its weight sharing nature.
		However, the estimated fitness is very noisy due to the co-adaptation of the
		operations in the supernet.	In this work, we propose a method called pEvoNAS
		wherein the whole neural architecture search space is progressively reduced to
		smaller search space regions with good architectures. This is achieved by
		using a trained supernet for architecture evaluation during the architecture
		search using genetic algorithm to find search space regions with good
		architectures. Upon reaching the final reduced search space, the supernet
		is then used to search for the best architecture in that search space using
		evolution. The search is also enhanced by using \textit{weight inheritance}
		wherein the supernet for the smaller search space inherits its weights
		from previous trained supernet for the bigger search space. Exerimentally,
		pEvoNAS gives better results on CIFAR-10 and CIFAR-100 while using
		significantly less computational resources as compared to previous EA-based
		methods. The code for our paper can be found
		\href{https://github.com/nightstorm0909/pEvoNAS}{here}.
	\end{abstract}
	
	\section{Introduction}
	\label{sec:intro}
	In the recent years, convolutional neural networks (CNNs) have been very
	instrumental in solving various computer vision problems. However, the CNN
	architectures (such as ResNet \cite{he2016deep}, DenseNet \cite{huang2017densely}
	AlexNet \cite{krizhevsky2012imagenet}, VGGNet \cite{simonyan2014very})
	have been designed mainly by humans, relying on their intuition and
	understanding of the specific problem. Searching the neural architecture
	automatically by using an algorithm, i.e. \textit{Neural architecture search}
	(NAS), is an alternative to the architectures designed by humans, and in the recent
	years, these NAS methods have attracted increasing interest because of its promise
	of an automatic and efficient search of architectures specific to a task.
	Vanilla NAS methods \cite{elsken2018neural}
	\cite{zoph2016neural}\cite{zoph2018learning} have shown promising results in the
	field of computer vision but most of these methods consume a huge amount of
	computational power as it involves training each architecture from scratch for its
	evaluation. Vanilla evolutionary algorithm (EA)-based NAS methods also suffers from
	the same huge computational requirement problem. For example, the method proposed
	in \cite{real2019regularized} required 3150 GPU days of evolution.
	\comm{Recently proposed gradient-based methods such as \cite{liu2018darts2}
		\cite{dong2019searching}\cite{xie2018snas}\cite{dong2019one}
		\cite{chen2019progressive} have reduced	the search time	by sharing weights among
		the architectures. However, these gradient-based methods highly depend on the
		given search space and suffer from premature convergence to the local optimum as
		shown in \cite{chen2019progressive} and \cite{Zela2020Understanding}.
	}
	
	\begin{figure}[t]
		\centering
		\begin{center}
			\includegraphics[width=0.9\linewidth]{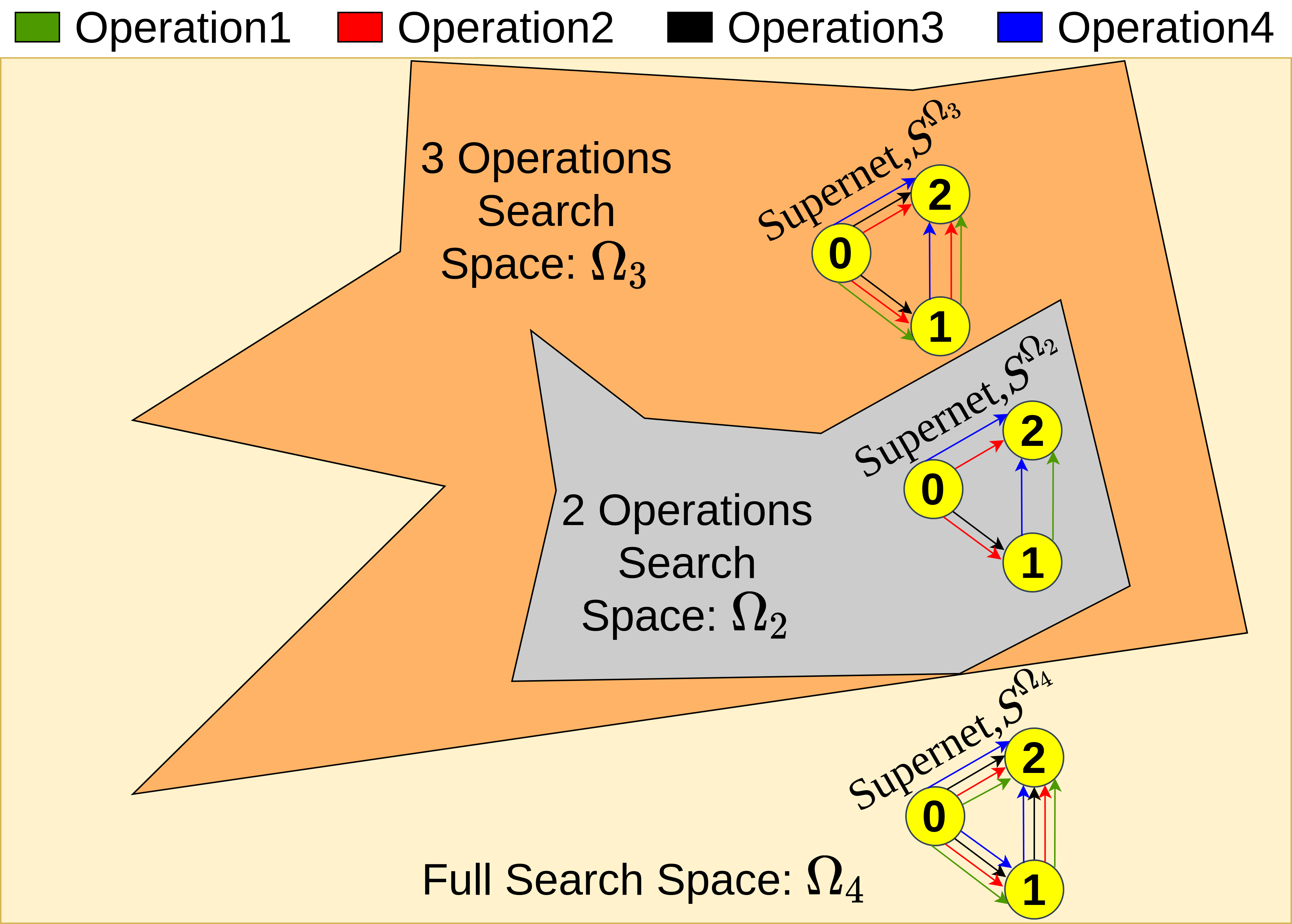}
		\end{center}
		\caption{Abstract illustration of progressive reduction of search space along
			with its corresponding supernet. The search starts from full search space
			($\Omega_4$) with 4 operations and is then progressively reduced to smaller
			search space regions: $\Omega_3 , \Omega_2$ with 3 and 2 operations
			respectively. Different colored arrows in supernets are used to represent
			different operations allowed in a specific search space.
		}
		\label{fig:search_space_progression}
	\end{figure}
	In this paper, we propose a method called pEvoNAS (\textit{Neural Architecture
		Search using Progressive Evolution}), which involves progressively reducing the
	search space by identifying the search space regions with good solution through
	a genetic algorithm. The \textit{fitness}/performance of an architecture in any
	search space is estimated using a supernet which is created using the allowed
	operations in a particular search space. A supernet represents all possible
	architectures in the search space while sharing the weights among all the
	architectures. As illustrated in Figure~\ref{fig:search_space_progression}, the
	architecture search	starts from the	full search space ($\Omega_4$) with 4
	operations for 3 nodes and a supernet, $S^{\Omega_{4}}$, is created with the 3
	nodes and 4 different colored arrows representing \textit{compound edges}
	(i.e. parallel operations) between the nodes. Reducing the search space to
	$\Omega_3$ and $\Omega_2$ involves reducing the number of operations between any
	two nodes to 3 and 2 respectively, which is reflected in their corresponding
	supernets, ($S^{\Omega_{3}}$, $S^{\Omega_{2}}$). The use of supernet for
	architecture evaluation results in reduction of search time as compared to
	vanilla EA-based methods. Also, it can be observed that the supernet for the
	larger search space (e.g. $S^{\Omega_{3}}$)	includes all the operations	present
	in the supernet for the smaller search space (e.g. $S^{\Omega_{2}}$). So, instead
	of intializing the smaller supernet, e.g. $S^{\Omega_{2}}$, with random weights, it
	inherits the respective weights from the previous bigger supernet,
	$S^{\Omega_{3}}$. This is known as \textit{supernet weight inheritance}.
	
	\comm{This is attributed to the evaluation process wherein each architecture
		is trained from scratch for a certain number of epochs in order to evaluate
		its performance on the validation data. Recently proposed gradient-based methods
		such as \cite{liu2018darts2}, \cite{dong2019searching}, \cite{xie2018snas},
		\cite{dong2019one}, \cite{chen2019progressive} have reduced the search time
		by sharing weights among the architectures. However, these gradient-based
		methods highly depend on the given search space and suffer from
		premature convergence to the local optimum as shown in
		\cite{chen2019progressive} and \cite{Zela2020Understanding}.
		
		\begin{figure}[t]
			
			\centering
			\begin{subfigure}{\linewidth}
				\includegraphics[width=\linewidth]{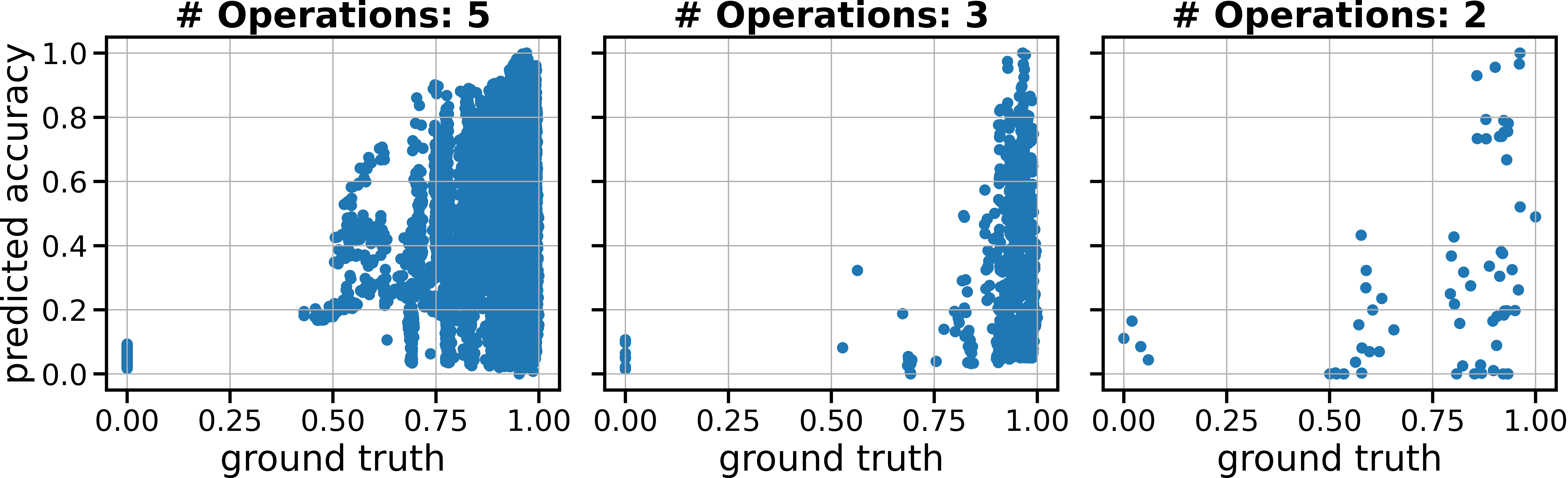}
				\caption{}
			\end{subfigure}
			\begin{subfigure}{\linewidth}
				\centering
				\begin{tabular}{|c|c|c|c|}
					\hline
					$\#$ Operations& 5 & 3 & 2 \\
					\hline
					Kendall Tau & $0.16$&$0.17$&$0.35$\\		
					\hline
				\end{tabular}
			\end{subfigure}
			\caption{
				Correlation score (Kendall Tau) at different number of operations.		
			}
			\label{fig:supernet_values}
		\end{figure}
	}
	
	\begin{table}[t]
		\centering
		\begin{tabular}{|c|c|c|c|}
			\hline
			$\#$ Operations& 5 & 3 & 2 \\
			\hline
			Kendall Tau & $0.16$&$0.17$&$0.35$\\		
			\hline
		\end{tabular}
		\caption{Correlation score (Kendall Tau) at different number of operations.
		}
		\label{table:kendall_tau}
	\end{table}
	
	Our contributions can be summarized as follows:
	\begin{itemize}
		\item We designed a framework of progressively reducing the search space
		to regions in the search space with good solutions using genetic algorithm.
		
		\item We also show the effectiveness of the \textit{weight inheritance for supernet}
		wherein a supernet for a smaller search space inherits weights from the trained
		supernet for the bigger search space.
		
		\item We also created a visualization of the progressive reduction of search
		space performed by pEvoNAS to get insights into the search process and show that 
		pEvoNAS indeed reduces the search space to good quality regions in the search space.
	\end{itemize}
	
	\section{Motivation}
	The use of supernet for architecture evaluations results in degraded architecture
	search performance because of
	the	inaccurate performance estimation by the supernet. This was first reported
	in \cite{bender2018understanding} and they showed that the co-adaptation among the
	operations in the compound edge leads to low correlation between the predicted
	performance via	supernet and the true architecture performance from
	training-from-scratch. In other words, the effect of co-adaptation is due to the
	combined operations in the compound edge. Following the logic, it seems reasonable
	that the supernet prediction will improve as number of operations reduce in the
	compound edge. For example, in Figure~\ref{fig:search_space_progression},
	the supernet $S^{\Omega_{3}}$ has 3 operations between any two nodes as compared
	to 2 operations in $S^{\Omega_{2}}$.
	
	We designed a controlled exeriment to test the assumption that \textit{as we
		reduce the number of operations in compound edge, the performance prediction
		of supernet improves}. We use NAS-Bench-201 \cite{Dong2020NAS-Bench-201} search
	space, which has 5 operations in the full search space. First, we train the
	supernet created using the operations in the full search space for 50 epochs and
	then compare the supernet predicted architecture performance of all the
	architectures in the search space with the ground truth provided in NAS-Bench-201.
	We randomly reduce the number of operations from 5 to 3 operations and create a
	supernet for the new smaller search space with weights inherited from previous
	trained supernet. We repeat the training of the new supernet again
	for 50 epochs and then comparing its prediction. Lastly, we repeat the process
	again for the new random search space of 2 operations.
	In Table~\ref{table:kendall_tau}, the correlation coefficient, Kendall's Tau
	\cite{kendall1938new}, is used to measure the correlation between estimated accuracy
	via supernet and ground truth accuracy and found that it increases as we
	reduce the number of operations considered between any two nodes in the
	supernet. Based on this observation, we use evolution to find smaller regions with good
	solutions while using supernet for the architecture evaluation in the search spaces. 
	
	\comm{Using this observation, we use the supernet to guide the search process
		using evolution to regions with good solutions until the final search space
		wherein the supernet is used to find a good architecture.}
	
	
	\section{Related Work}
	\comm{Early NAS approaches\cite{stanley2002evolving}\cite{stanley2009hypercube},
		optimized both the neural architectures and the weights of the network using
		evolution. However,	their usage was limited to shallow networks. Recent
		\cite{zoph2016neural}\cite{pmlr-v80-pham18a}\cite{real2019regularized}
		\cite{zoph2018learning}\cite{real2017large}\cite{liu2018hierarchical}
		\cite{xie2017genetic} perform the architecture search separately while using
		gradient descent for optimizing the weights of the architecture for its
		evaluation which has made the search of deep networks possible.}
	The various NAS
	methods can be classified into two categories: \textit{gradient-based} methods
	and \textit{non-gradient based} methods.
	
	\textbf{Gradient-Based Methods:} \comm{These methods begin with a random neural
		architecture and the neural architecture is then updated using the gradient
		information on the basis of its performance on the validation data.}
	In general,	these methods, \cite{liu2018darts2}\cite{dong2019searching}
	\cite{xie2018snas}\cite{dong2019one}, relax the discrete architecture search
	space to a continuous search space by using a supernet. The	performance of the
	supernet on the validation data is used for updating the architecture using
	gradients. As the supernet shares weights among all architectures
	in the search space, these methods take	lesser time in the evaluation process
	and thus shorter search time. However, these methods suffer from the overfitting
	problem	wherein the	resultant architecture shows good performance on the validation
	data but exhibits poor performance on the test data. This can be attributed to its
	preference for parameter-less operations in the search space, as it leads to rapid
	gradient descent, \cite{chen2019progressive}.
	In contrast to these gradient-based methods, our method does not suffer from the
	overfitting problem because of its stochastic nature.
	
	\textbf{Non-Gradient Based Methods:}
	These methods include reinforcement learning (RL) methods and evolutionary
	algorithm (EA) methods. In the RL methods \cite{zoph2016neural}
	\cite{zoph2018learning}, an agent is used for the generating neural architecture
	and the agent is then trained to generate architectures in order to maximize
	its	expected accuracy on the validation	data. These accuracies were calculated by
	training the architectures from	scratch to convergence which resulted in long
	search time. This was improved in \cite{pmlr-v80-pham18a} by using a single
	directed acyclic graph (DAG) for sharing the weights among all the sampled
	architectures, thus resulting in reduced computational resources.
	The EA based NAS methods begin with a population of architectures and each
	architecture in the population is evaluated on the basis of its performance on the
	validation data. The popluation is then evolved	on the basis of the performance of
	the population. Methods such as those proposed in \cite{real2019regularized} and
	\cite{xie2017genetic} used gradient descent for	optimizing the weights of each
	architecture in the population from scratch in order to determine their accuracies 
	on the validation data as their fitness, resulting in huge computational
	requirements. In order to speed up the training process, in \cite{real2017large},
	the authors introduced weight inheritance wherein the architectures in the
	new generation population inherit the weights of the previous generation
	population, resulting in bypassing the training from scratch. However, the speed
	up gained is less as it still needs	to optimize the weights of the architecture.
	Methods such as that proposed in \cite{sun2019surrogate} used a random forest for
	predicting the performance of the architecture during the evaluation process,
	resulting in a high speed up as	compared to previous EA methods. However, its
	performance was far from the state-of-the-art results. In contrast, our method
	achieved better results than previous EA methods while using significantly less
	computational resources.
	
	\begin{figure}[h]
		\centering
		\begin{center}
			\includegraphics[width=\linewidth]{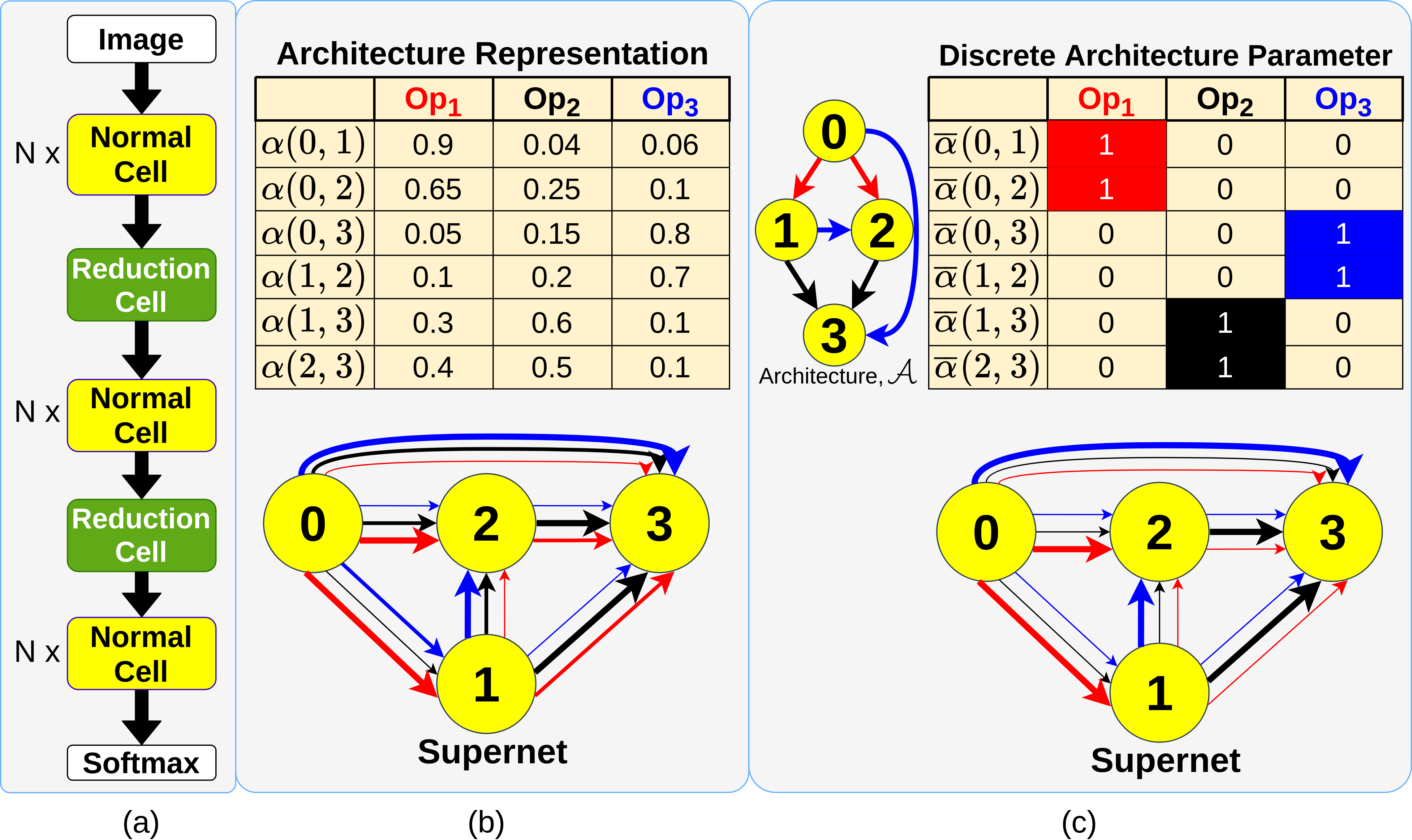}
		\end{center}
		\caption{
			(a) Architecture created by staking cells.
			(b) Architecture representation for a search space with 3 operations
			and 4 nodes. The thickness of the arrow in the supernet is proportional
			to the weight given to an operation.
			(c) Illustration of selecting an architecture in the supernet. The
			highlighted cell represents the selected operation between any two nodes.
		}
		\label{fig:arch_representation}
	\end{figure}
	\section{Proposed Method}
	\subsection{Search Space and Architecture Representation}
	\label{subsect:search_space}	
	Following \cite{pmlr-v80-pham18a}\cite{real2019regularized} \cite{zoph2018learning}
	\cite{liu2018darts2}\cite{dong2019searching}\cite{dong2019one}\cite{lu2020multi}
	\cite{liu2018progressive}, the architecture is created by staking together cells
	of two types: \textit{normal} cells which preserve the dimentionality of the
	input with a stride of one and \textit{reduction} cells which reduce the spatial
	dimension with a stride	of two, shown in Figure~\ref{fig:arch_representation}(a).
	\comm{
		In this	work, we applied our method to two different search spaces: \textit{Search
			space 1	(\textbf{S1})},	(\cite{liu2018darts2}) and \textit{Search space 2
			(\textbf{S2})},	(\cite{Dong2020NAS-Bench-201}). In \textit{S1}, we search for both
		normal and reduction cells where each node	$x^{(j)}$ maps two inputs to one
		output. In \textit{S2}, we search for only normal cells, where each node $x^{(j)}$
		is connected to the previous node $x^{(i)}$ (i.e. $i < j$).
	}
	As illustrated in Figure~\ref{fig:arch_representation}(b), a cell in the
	architecture is	represented	by an \textit{architecture parameter}, $\alpha$.
	Each $\alpha$ for a	normal cell and a reduction cell is represented by a matrix
	with columns representing the weights of different operations $Op(.)s$ from the
	operation space	$\mathcal O$ (i.e. the search space of NAS) and rows representing
	the edge between two nodes. For example, in
	Figure~\ref{fig:arch_representation}(b), $\alpha(0, 1)$ represents the edge
	between node 0 and node 1 and the entries in the row represent the weights given
	to the three different operations. 
	
	\subsection{Performance Estimation}
	\label{subsect:performance}
	We used a \textit{supernet} \cite{liu2018darts2} to estimate the performance of
	an architecture in the search space. It shares the weights among all architectures
	in the search space	by treating all the architectures as the subgraphs
	of a supergraph. As illustrated in Figure~\ref{fig:arch_representation}(b), the
	supernet uses the \textit{architecture parameter}, $\alpha$, by normalizing it
	using softmax. The directed edge from node $i$ to node $j$ is the weighted sum of
	all $Op(.)s$ in $\mathcal O$ where the $Op(.)s$ are	weighted by the normalized
	$\alpha^{(i,j)}$. This can be written as:
	\begin{equation}
		\label{eq:ons}
		f^{(i,j)}(x^{(i)}) = \sum_{Op \in \mathcal O } 
		\frac{exp(\alpha^{(i,j)}_{Op})}
		{\sum_{Op' \in \mathcal O } exp(\alpha^{(i,j)}_{Op'})} Op(x^{(i)})
	\end{equation}
	where $\alpha^{(i,j)}_{op}$ represents the weight of the operation $Op(.)$ in
	the operation space $\mathcal O$ between node $i$ and node $j$. This design
	choice allows us to skip the individual architecture training from
	scratch for its evaluation because of the weight-sharing nature of the supernet,
	thus resulting in a significant reduction of search time. The supernet is trained
	on training dataset for a certain number of epochs using \textit{Stochastic
		Gradient Descent} (SGD), \cite{sutskever2013importance} with momentum.
	During training, random architecture parameter, $\alpha$, is sent to the supernet
	for each training batch in an epoch so that no particular sub-graph (i.e.
	architecture) of the super-graph (i.e. supernet) receives most of the gradient
	updates.
	
	The performance	of an architecture is calculated using the trained supernet on
	the validation data, also known as the \textit{fitness} of the architecture. As
	illustrated in Figure~\ref{fig:arch_representation}(c), in order to select an
	architecture, $\mathcal A$, in the supernet, a new architecture parameter called
	\textit{discrete architecture} parameter, $\bar\alpha$ is created with the
	following entries:
	\begin{equation}
		\bar{\alpha}^{(i,j)}_{Op} = 
		\begin{cases}
			1, \text{if $Op(x^{(i)})$ present in $A$}\\
			0,  \text{otherwise}\\
		\end{cases}    
	\end{equation}
	Using $\bar\alpha$, the architecture, $A$, is selected in the supernet
	and the accuracy of the supernet on the validation data is used as the estimated
	\textit{fitness} of $A$.
	\comm{This process gives higher equal
		weight to the architecture $A$ operations while giving lower equal
		weights to the other architecture operations. This results in the
		higher contribution from the architecture $A$ while very low
		contribution by other architectures during the fitness evaluation.
		This process can be thought of as selecting a subgraph (i.e. $A$)
		from the supergraph	(i.e. supernet).
	}
	
	\begin{figure}[b]
		\centering
		\begin{center}
			\includegraphics[width=\linewidth]{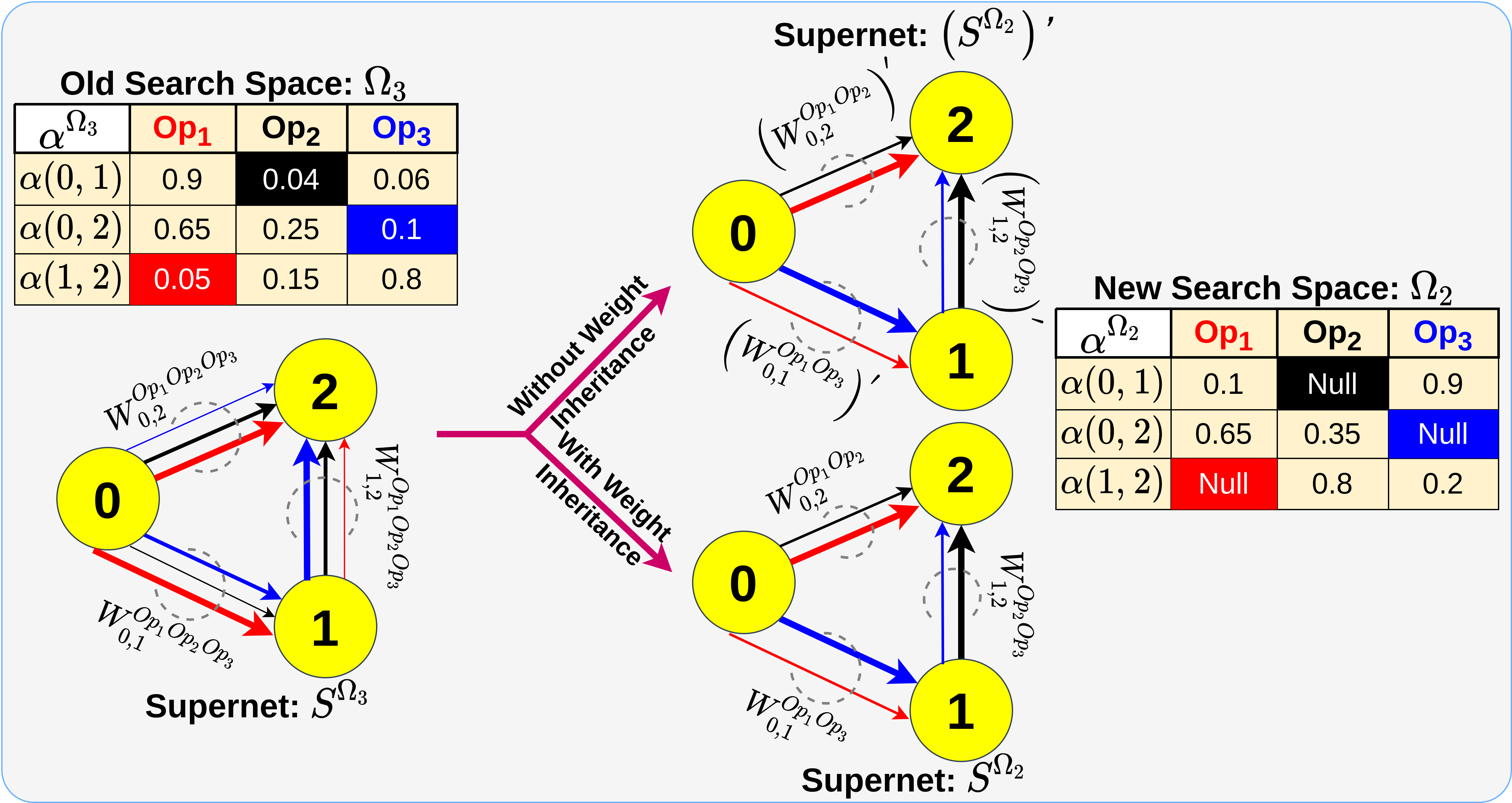}
		\end{center}
		\caption{Abstract illustration of reduction of search space from a bigger
			search space with 3 operations to smaller search space with 2
			operations. The highlighted cells are the operations that have been
			removed from the search space. $W_{n_1, n_2}^{Op(.)}$ represents the
			weights of the allowed operations between node $n_1$ and node $n_2$.
			$(W_{n_1, n_2}^{Op(.)})'$ represents randomly initialized weights.
		}
		\label{fig:search_space_reduce}
	\end{figure}
	
	\subsection{Search Space Reduction}
	\label{subsect:search_space_reduce}
	Reducing the search space involves reducing the number of operations considered
	between any two nodes. This is done by selecting the top-k operations in each row
	of the $\alpha$ where k represents the number of operations for the new reduced
	search space. For example, in Figure~\ref{fig:search_space_reduce}, the top-2
	operations ($Op_2, Op_3$) are selected in $\alpha(1,2)$ from the old search space,
	$\Omega_{3}$, to create a new search space $\Omega_{2}$. Since all the operations
	present in $\Omega_{2}$ are also present in $\Omega_{3}$, the supernet for
	$\Omega_{2}$, (i.e. $S^{\Omega_{2}}$), can be created in 2 ways: (i)
	\textbf{With Supernet Weight Inheritance}: Here, the smaller search space supernet
	inherits/copies the weights from the previous bigger search space supernet.
	(ii) \textbf{Without Supernet Weight Inheritance}: Here, the smaller search space
	supernet is created with random weights. For example, in
	Figure~\ref{fig:search_space_reduce}, the weights of the operations	between node 1
	and	2 for $S^{\Omega_{2}}$ are inherited from $S^{\Omega_{3}}$ while the weights
	of the operations in $(S^{\Omega_{2}})'$ are randomly intialized.
	
	\begin{algorithm}[h]
		\caption{pEvoNAS}	
		\label{algo:pEvoNAS}
		\SetAlgoLined
		\KwIn{Population size $N_{pop}$, 
			training data $\mathfrak{D}_{tr}$, validation data $\mathfrak{D}_{va}$,
			training epochs $N_{epochs}$, list of operation numbers
			$\mathfrak{Op}$, convergence number $N_{conv}$.
		}
		\KwOut{Best architecture, $E_{best}$.}
		\ForEach{ $op \in \mathfrak{Op}$}{%
			\uIf{smaller search space}{
					\tcc{Search space reduction}
					Reduce search space to $op$ operations, $\Omega_{op}$, using
					$E_{best}$\;
					\tcc{Weight inheritance}
					Create supernet, $S^{\Omega_{op}}$, with weights $W$ inherited
					from previous supernet\;
				}
				\Else(\tcp*[h]{For full search space}){
					Initialize supernet, $S^{\Omega_{op}}$, with random weights $W$ for
					full search space\;
				}
				
				TrainSupernet($S^{\Omega_{op}}$,$N_{epochs}$, $\mathfrak{D}_{tr}$)\;
				
				\tcc{Architecture search using evolution}
				$E_{best} \gets$ EA($\Omega_{op}$, $S^{\Omega_{op}}$, $N_{conv}$, $\mathfrak{D}_{va}$, $N_{pop}$)\;
				\If{Final search space}{
					\Return $E_{best}$
				}			
			}
		\end{algorithm}
		
		\subsection{pEvoNAS}
		\label{subsect:pEvoNAS}
		The entire process is summarized in	Algorithm~\ref{algo:pEvoNAS}. It begins
		with a list of number of operations, $\mathfrak{Op}$, to which the search space
		is to be reduced. For each operation number, $op \in \mathfrak{Op}$, a supernet,
		$S^{\Omega_{op}}$, is created for the search space, $\Omega_{op}$. If it is the
		first search space (i.e. full search space) then the weights of $S^{\Omega_{op}}$
		are randomly initialized, otherwise they are inherited from the previous search
		space trained supernet. The supernet, $S^{\Omega_{op}}$, is then trained for
		$N_{epochs}$ epochs. The evolutionary algorithm (EA) then performs the
		architecture search in $\Omega_{op}$. It starts with a population of $N_{pop}$
		architectures, which are sampled from a uniform distribution on the interval
		$\left[0,1\right)$.	\textit{Fitness} of each individual in the population is
		evaluated using the	trained $S^{\Omega_{op}}$. The population is then evolved
		using crossover and	mutation operations to create the next generation population
		replacing the previous generation population. The best architecture, $E_{best}$,
		in each generation does not undergo any modification and is automatically copied
		to	the	next generation. This ensures that the algorithm does not forget the best
		architecture learned thus far and gives an opportunity to old generation
		architecture to compete against the	new	generation architecture.
		The process of population evaluation and then evolution is
		repeated until $E_{best}$ does not change for $N_{conv}$ generations, showing the
		\textit{convergence} to an architecture. $E_{best}$ is then used to reduce the
		reduce the search space. The whole process is repeated until the search space is
		reduced to the final search	space, $\Omega_{final}$. For $\Omega_{final}$,
		$E_{best}$ is returned as the searched architecture. The pseudocodes for the
		supernet training and the evolutionary algorithm are given in the supplementary.
		\begin{figure}[h]
			\centering
			\begin{center}
				\includegraphics[width=0.65\linewidth]{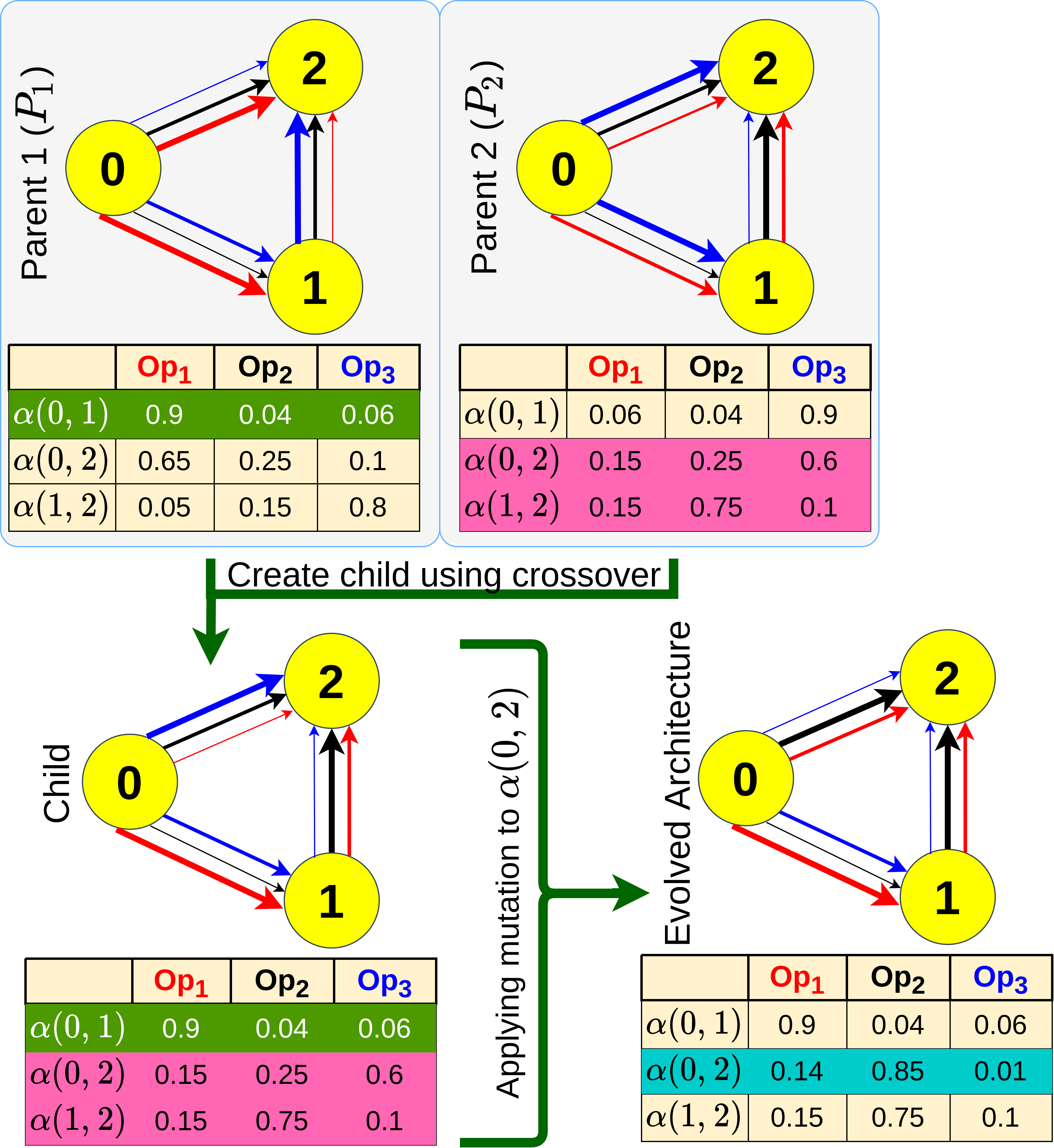}
			\end{center}
			\caption{Abstract illustration of the evolution process using crossover and
				mutation.
			}
			\label{fig:evolution_process}
		\end{figure}
		
		\textbf{Crossover and Mutation Operations}:
		Crossover combines 2 parents ($P_1$, $P_2$), selected through \textit{tournament
			selection} \cite{eiben2003introduction}, to create a new child architecture, which
		may	perform better than the parents. In	\textit{tournament selection}, a certain
		number of architectures are randomly selected from the current population and the
		most fit architecture from the selected group becomes the parent. We get 
		$P_1$ and $P_1$ on applying	tournament selection two times which are then used to
		create a single child architecture.	This is done by copying the edge
		between $node_i$ and $node_j$, from either $P_1$ or $P_2$, with 50\% probability,
		to the child architecture
		between $node_i$ and $node_j$.
		\comm{
			This can be	formulated as follows:
			\begin{equation}
				[{\alpha}^{i,j}]_{child} = 
				\begin{cases}
					[{\alpha}^{i,j}]_{P_1}, \text{with probability 0.5}
					\\
					[{\alpha}^{i,j}]_{P_2},  \text{otherwise}
					\\
				\end{cases}    
			\end{equation}
		}
		Mutation refers to a random change to an individual architecture in the population.
		The	algorithm uses the \textit{mutation rate} \cite{eiben2003introduction}, which
		decides the probability of changing the architecture parameter, $\alpha^{i,j}$,
		between	node $i$ and node $j$. This is done by re-sampling $\alpha^{i,j}$ from a 
		uniform distribution on the interval $\left[0,1\right)$. As illustrated in
		Figure ~\ref{fig:evolution_process}, ${\alpha}^{0,1}_{child}$ and
		${\alpha}^{1,2}_{child}$ are copied from $P_1$ and $P_2$ respectively during
		crossover while applying mutation to ${\alpha}^{0,2}_{child}$ in the child
		architecture.
		
		\section{Experiments}
		\label{experiments}
		\subsection{Search Spaces}
		In this section, we report the performance of progEvoNAS in terms of a neural 
		architecture search on two different search spaces: 1) \textit{Search
			space 1	(\textbf{S1})} \cite{liu2018darts2} and 2) \textit{Search space 2
			(\textbf{S2})}\cite{Dong2020NAS-Bench-201}. In \textit{S1}, we search for both
		normal and reduction cells where each node	$x^{(j)}$ maps two inputs to one
		output. Here, each cell has seven nodes with first two nodes being the output from
		previous cells and last node as output node, resulting in 14 edges among them.
		There are 8 operation in S1, so each architecture is represented by two 14x8
		matrices, one each for normal cell and reduction cell. In \textit{S2}, we search for
		only normal cells, where each node $x^{(j)}$ is connected to the previous node
		$x^{(i)}$ (i.e. $i < j$). It is a smaller search space where we only search for the
		normal cell in Figure~\ref{fig:arch_representation}(a). It provides a unified
		benchmark for almost any up-to-date NAS algorithm by providing results of each
		architecture in the search space on	CIFAR-10, CIFAR-100 and ImageNet-16-12. Here,
		each cell has four nodes with first node as input node and last node as output node,
		resulting in 6 edges among them. There are 5 operations in S2, so each architecture
		is represented by one 6x5 matrix for the normal cell.
		
		\subsection{Dataset} \textbf{CIFAR-10} and \textbf{CIFAR-100}
		\cite{krizhevsky2009learning} has 50K training images and 10K testing images with
		images classified into 10 classes and 100 classes respectively. \textbf{ImageNet}
		\cite{imagenet_cvpr09} is well known benchmark for image classification
		containing 1K classes with 1.28 million training images and 50K images test images.
		\textbf{ImageNet-16-120} \cite{chrabaszcz2017downsampled} is a down-sampled variant
		of ImageNet where the original ImageNet is downsampled to 16x16 pixels with labels
		$\in\left[0,120\right]$ to construct ImageNet-16-120 dataset.
		The settings used for the datasets in \textbf{S1} are as follows:
		\begin{itemize}
			\item \textit{CIFAR-10}: We split 50K training images into two sets of size
			25K each, with one set acting as the training set and the other set as the
			validation set.
			\item \textit{CIFAR-100}: We split 50K training images into two sets. One set
			of size 40K images becomes the training set and the other set of size 10K
			images becomes the validation set.
		\end{itemize}
		The settings used for the datasets in \textbf{S2} are as follows:
		\begin{itemize}
			\item \textit{CIFAR-10}: The same settings as those used for S1 is used here
			as well.
			\item \textit{CIFAR-100}: The 50K training images remains as the training set
			and the 10K testing images are split into two sets of size 5K each, with one
			set acting as the validation set and the other set as the test set.
			\item \textit{ImageNet-16-120}: It has 151.7K training images, 3K validation
			images and 3K test images.
		\end{itemize}
		
		\subsection{Implementation Details}
		\label{implementation_details}
		
		\subsubsection{\textbf{Supernet Training Settings:}}
\begin{figure}[t]
			\centering
			\begin{center}
				\includegraphics[width=\linewidth]{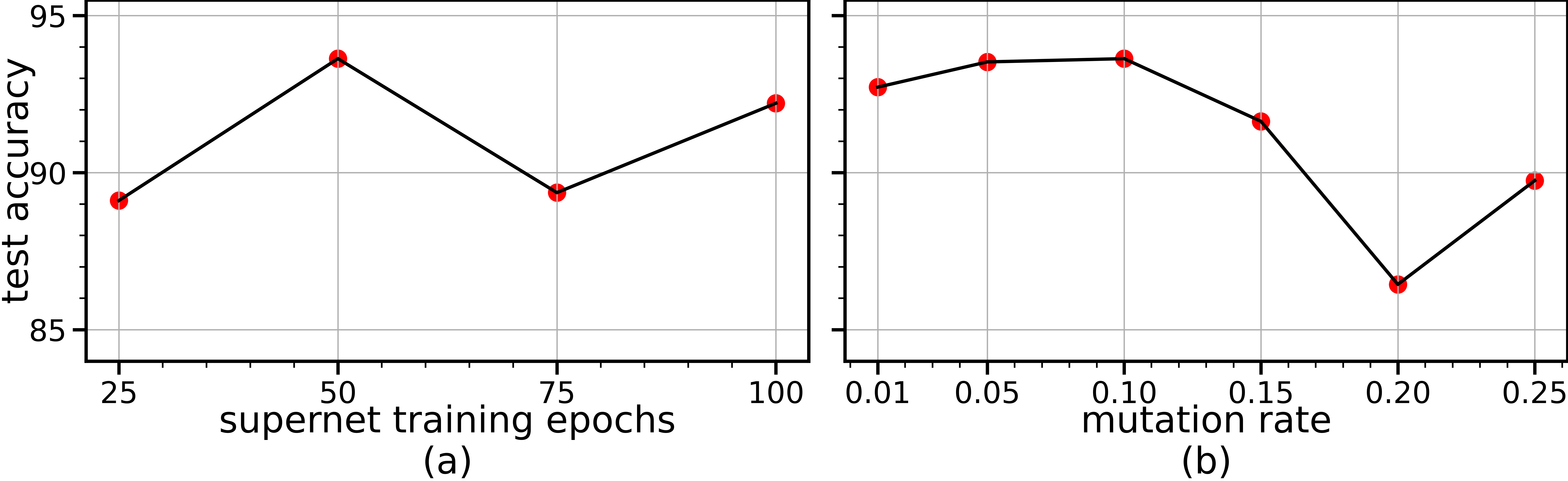}
			\end{center}
			\caption{Tests conducted for hyperparameter selection in S2.
				Test accuracy vs (a) training epochs of supernet, and
				(b) mutation rate.
			}
			\label{fig:hyperparameter}
		\end{figure}

		In general, the supernet suffers from high memory requirements which makes it
		difficult to fit it in a single GPU. For S1, we follow \cite{liu2018darts2}
		and use a smaller supernet, called \textit{proxy model} which is created with 20
		stacked cells and 16 initial channels. It is trained on both CIFAR-10 and
		CIFAR-100 with SGD for 50 epochs (i.e. $N_{epochs}$), which is chosen based on
		the experiment conducted in S2, shown in Figure~\ref{fig:hyperparameter}(a).
		All the other settings are also same for both datasets i.e. batch size of 64,
		weight 	decay $\lambda=3\times10^{-4}$, cutout\cite{devries2017improved},
		initial learning rate $\eta_{max}=0.025$ (annealed down to 0 by using a cosine
		schedule without restart\cite{DBLP:conf/iclr/LoshchilovH17}) and momentum
		$\rho=0.9$. For	\textit{S2}, we do not use a proxy model as the size of the
		supernet is sufficiently small to be fitted in a single GPU. For training, we
		follow the same settings as those used in S1 for CIFAR-10, CIFAR-100 and
		ImageNet16-120 except batch size of 256.
		
		\subsubsection{\textbf{Architecture evaluation:}} Here, the discovered architecture,
		$E_{best}$ (i.e. discovered cells), at the end of the architecture search is
		trained on the dataset to evaluate its performance for comparing with other
		NAS methods. For \textit{S1}, we follow the training settings used in DARTS
		\cite{liu2018darts2}. Here, a larger network, called \textit{proxyless network} 
		\cite{li2019random}, is created using $E_{best}$ with 20 stacked cells and
		36 initial channels for both CIFAR-10 and CIFAR-100 datasets. It is then
		trained for 600 epochs on both the datasets with the same settings as the ones
		used in the supernet training above. Following recent works
		\cite{pmlr-v80-pham18a}\cite{real2019regularized}\cite{zoph2018learning}
		\cite{liu2018darts2}\cite{liu2018progressive}, we use an auxiliary tower with
		0.4 as its	weights, path dropout probability of 0.2 and cutout
		\cite{devries2017improved} for additional enhancements.	For	ImageNet, $E_{best}$ is
		created with 14 cells and 48 initial channels in the mobile setting, wherein the input
		image size is 224 x 224 and the number of multiply-add operations in the model is
		restricted to less than 600M. It is trained on 8 NVIDIA V100 GPUs by following the
		training settings used in \cite{chen2019progressive}.
		
		\begin{table}[t]
			\caption{Comparison of pEvoNAS with other NAS methods in S1 in terms
				of test accuracy (higher is better) on CIFAR-10.}
			\label{table:c10_s1}
			\begin{subtable}{\textwidth}
				\scalebox{0.85}{
					\begin{tabular}{l|c|c|c|c}
						
						\hline
						\multicolumn{1}{c|}{\bf{Architecture}} & \multicolumn{1}{c|}{\bf{Top-1}} & \bf{Params} & \bf{GPU} &\bf{Search} \\
						&\textbf{Acc.} (\%)&(M)& \bf{Days} & \bf{Method}\\
						\hline
						ResNet\cite{he2016deep}  & 95.39 & 1.7 & - & manual\\
						DenseNet-BC\cite{huang2017densely}& 96.54 & 25.6 & - & manual\\
						ShuffleNet\cite{zhang2018shufflenet}& 90.87 & 1.06 & - & manual\\
						\hline
						
						PNAS\cite{liu2018progressive}  & 96.59 & 3.2 & 225 & SMBO\\
						RSPS\cite{li2019random}         & 97.14 & 4.3 &2.7 & random\\
						\hline
						
						NASNet-A\cite{zoph2018learning}     & 97.35 & 3.3  &1800 & RL\\
						ENAS\cite{pmlr-v80-pham18a}    & 97.14  & 4.6  &0.45 & RL\\
						\hline
						DARTS\cite{liu2018darts2}     & 97.24 & 3.3 &4 & gradient\\
						GDAS\cite{dong2019searching}& 97.07 & 3.4 &0.83 & gradient\\
						SNAS\cite{xie2018snas}       & 97.15 & 2.8&1.5 & gradient\\
						SETN\cite{dong2019one}       & 97.31 & 4.6 &1.8 & gradient\\
						\hline
						
						AmoebaNet-A\cite{real2019regularized} & 96.66 & 3.2 &3150& EA\\
						Large-scale Evo.\cite{real2017large} & 94.60 & 5.4 &2750 & EA\\
						Hierarchical Evo.\cite{liu2018hierarchical}  & 96.25 & 15.7 &300 & EA\\
						CNN-GA\cite{sun2020automatically}  & 96.78 & 2.9 &35 & EA\\
						CGP-CNN\cite{suganuma2017genetic}  & 94.02 & 1.7 &27 & EA\\
						AE-CNN\cite{sun2019completely}  & 95.7 & 2.0 &27 & EA\\
						NSGANetV1-A2\cite{lu2020multi}  & 97.35 & 0.9 &27& EA\\
						AE-CNN+E2EPP\cite{sun2019surrogate}  & 94.70 & 4.3 &7 & EA\\
						NSGA-NET\cite{lu2019nsga}  & 97.25 & 3.3 &4& EA\\
						\hline
						\bf{pEvoNAS-C10A}   & \bf{97.52} & \bf{3.6} & \bf{1.20}&\bf{EA}\\
						pEvoNAS-C10B              & 97.36      & 3.5  & 1.31  & EA\\
						pEvoNAS-C10C              & 97.27      & 3.0  & 1.41  & EA\\
						\hline
						pEvoNAS-C10rand           & 96.83   & 3.37 & 0.11  & random\\
						\hline
					\end{tabular}
				}
			\end{subtable}
		\end{table}
		
		\begin{table}[h]
			\caption{Comparison of pEvoNAS with other NAS methods in S1 in terms
				of test accuracy (higher is better) on CIFAR-100.}
			\label{table:c100_s1}
			\begin{subtable}{\textwidth}
				\scalebox{0.85}{
					\begin{tabular}{l|c|c|c|c}
						\hline
						\multicolumn{1}{c|}{\bf{Architecture}} &  \multicolumn{1}{c|}{\bf{Top-1}} & \bf{Params} & \bf{GPU} &\bf{Search} \\
						&\textbf{Acc.} (\%)& (M) & \bf{Days} &\bf{Method}\\
						\hline
						ResNet\cite{he2016deep}  & 77.90 & 1.7 & - & manual\\
						DenseNet-BC\cite{huang2017densely} & 82.82   & 25.6 & - & manual\\
						ShuffleNet\cite{zhang2018shufflenet}& 77.14 & 1.06 & - & manual\\
						\hline
						PNAS\cite{liu2018progressive}  & 80.47 & 3.2 & 225 & SMBO\\
						\hline
						
						\comm{NASNet-A\cite{zoph2018learning}    & -     & 3.3  &1800& RL\\}
						MetaQNN\cite{baker2017designing}     & 72.86      & 11.2  &90& RL\\
						ENAS\cite{pmlr-v80-pham18a}     & 80.57      & 4.6  &0.45& RL\\
						\hline
						
						
						AmoebaNet-A\cite{real2019regularized}  & 81.07 & 3.2 &3150& EA\\
						Large-scale Evo.\cite{real2017large}  & 77.00 & 40.4 &2750& EA\\
						CNN-GA\cite{sun2020automatically}  & 79.47 & 4.1 &40& EA\\
						AE-CNN\cite{sun2019completely}  & 79.15 & 5.4 &36& EA\\
						NSGANetV1-A2\cite{lu2020multi} & 82.58 & 0.9 &27& EA\\
						Genetic CNN\cite{xie2017genetic}  & 70.95 & - &17& EA\\
						AE-CNN+E2EPP\cite{sun2019surrogate}  & 77.98 & 20.9 &10& EA\\
						NSGA-NET\cite{lu2019nsga}  & 79.26 & 3.3 &8& EA\\
						
						\hline
						\bf{pEvoNAS-C100A}    & \bf{82.59} & \bf{3.0} & \bf{1.25} & \bf{EA}\\
						pEvoNAS-C100B         & 82.44   & 3.1      & 1.28 & EA\\
						pEvoNAS-C100C         & 82.23   & 3.3      & 1.22 & EA\\
						\hline
						pEvoNAS-C100rand      & 81.03   & 2.8 & 0.15 & random\\
						\hline
					\end{tabular}
				}
			\end{subtable}
		\end{table}
		
		\begin{table}[h]
			\caption{Comparison of pEvoNAS with other NAS methods in S1 in terms
				of test accuracy (higher is better) on ImageNet.}
			\label{table:imagenet_s1}
			\begin{subtable}{\textwidth}
				\scalebox{0.67}{
					\begin{tabular}{l|c|c|c|c|c|c}
						\hline
						\multicolumn{1}{c|}{\bf{Architecture}} & \multicolumn{2}{c|}{\bf{Test Accuracy (\%)}} & \bf{Params} &+$\times$& \bf{GPU} &\bf{Search} \\
						& \bf{top 1} & \bf{top 5} & (M) & (M) & \bf{Days} & \bf{Method} \\
						
						\hline
						MobileNet-V2, (\cite{sandler2018mobilenetv2})&72.0& 91.0 & 3.4 & 300 & - & manual\\ 
						\hline
						PNAS, (\cite{liu2018progressive})  &74.2& 91.9    & 5.1 & 588 & 225 & SMBO\\
						\hline
						
						NASNet-A, (\cite{zoph2018learning})    & 74.0 & 91.6      & 5.3 & 564 &1800& RL\\
						NASNet-B, (\cite{zoph2018learning})    & 72.8  & 91.3    & 5.3 & 488 &1800& RL\\
						NASNet-C, (\cite{zoph2018learning})    & 72.5  & 91.0     & 4.9 & 558 &1800& RL\\
						\hline
						
						DARTS, (\cite{liu2018darts2})& 73.3  & 91.3 & 4.7 & 574 &4& gradient\\
						GDAS, (\cite{dong2019searching})  & 74.0  & 91.5 & 5.3 & 581 &0.83& gradient\\
						SNAS, (\cite{xie2018snas})& 72.7  & 90.8 & 4.3 & 522 &1.5& gradient\\
						SETN, (\cite{dong2019one})& 74.3  & 92.0 & 5.4 & 599 &1.8& gradient\\
						\hline
						
						AmoebaNet-A, (\cite{real2019regularized}) & 74.5 & 92.0 & 5.1 & 555 &3150& EA\\
						AmoebaNet-B, (\cite{real2019regularized}) & 74.0 & 91.5 & 5.3 & 555 &3150& EA\\
						AmoebaNet-C, (\cite{real2019regularized}) & 75.7 & 92.4 & 6.4 & 570 &3150& EA\\
						NSGANetV1-A2, (\cite{lu2020multi}) & 74.5 & 92.0 & 4.1 & 466 &27& EA\\
						\comm{FairNAS-A          \cite{chu2019fairnas} & 75.3 & 92.4 & 4.6 & 388 &12& EA\\
							FairNAS-B          \cite{chu2019fairnas} & 75.1 & 92.3 & 4.5 & 345 &12& EA\\
							FairNAS-C          \cite{chu2019fairnas} & 74.7 & 92.1 & 4.4 & 321 &12& EA\\
							\hline
							
							EvNAS-A (Ours)                                     & 75.6 & 92.6 & 5.1 & 570 & 3.83 & EA\\
							EvNAS-B (Ours)                                     & 75.6 & 92.6 & 5.3 & 599 & 3.83 & EA\\
							EvNAS-C (Ours)                                     & 74.9 & 92.2 & 4.9 & 547 & 3.83 & EA\\
						}
						\hline
						\bf{pEvoNAS-C10A} & \bf{74.9} & \bf{92.4} & \bf{5.1} & \bf{567} & \bf{1.20} & \bf{EA}\\
						pEvoNAS-C100A     & 73.2 & 91.3 & 4.3 & 478 & 1.25 & EA\\
						\hline
					\end{tabular}
				}
			\end{subtable}
		\end{table}

		\subsubsection{\textbf{Evolutionary Algorithm Settings:}}
		The architecture search begins with the full search	space for both the search
		spaces. So, for \textit{S1}, the architecture search begins with the search space
		with 8 operations which is then progressively reduced to smaller search space
		regions	with 5 operations and then finally to 2 operations. While, for \textit{S2},
		the architecture search begins with the search space with 5 operations which is
		then progressively reduced to search spaces with 3 and 2 operations. Following
		\cite{sun2019surrogate}\cite{sun2019completely}, the evolutionary
		algorithm (EA), for both \textit{S1} and \textit{S2}, uses a population size of 20
		in each	generation. For the tournament selection, 5 architectures are chosen randomly
		from the current population and the best best architecture among them becomes the
		parent. We apply the tournament selection 2 times to get 2 parents for the crossover
		operation. \textit{Mutation rate} of 0.1 was chosen based on the experiment	conducted in
		S2, shown in Figure~\ref{fig:hyperparameter}(b). The evolutionary search runs until the
		best architecture, $E_{best}$, is repeated for 10 generations (i.e. $N_{conv}$).
		All the above training and architecture search were performed on a single Nvidia
		RTX 3090 GPU.
		
		\begin{figure}[b]
			\centering
			\begin{center}
				\includegraphics[width=\linewidth]{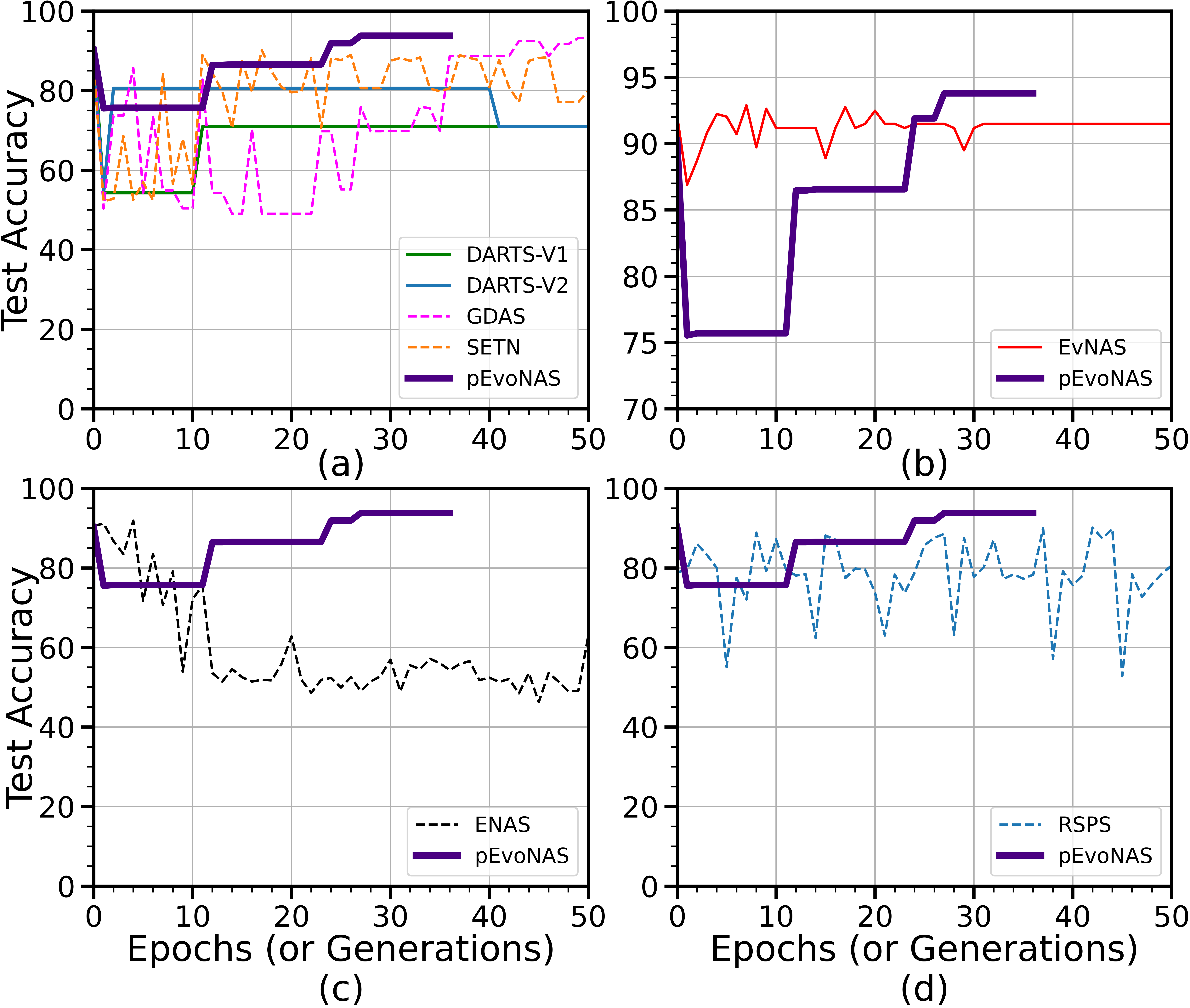}
			\end{center}
			\caption{Comparision of pEvoNAS with
				(a) gradient-based methods,
				(b) EA-based method,
				(c) RL method and
				(d) random search
				on CIFAR-10 for the search space S2.
			}
			\label{fig:s2_comparison}
		\end{figure}
		
		\begin{table*}[t]
			\caption{Comparison of pEvoNAS with other NAS methods on NAS-Bench-201 (i.e. 
				S2)\cite{Dong2020NAS-Bench-201} with mean $\pm$ std. accuracies on
				CIFAR-10, CIFAR-100 and ImageNet16-120 (higher is better).
				Optimal refers to the best architecture accuracy for each dataset.
				Search times are given for a CIFAR-10 search on a single GPU.}
			\label{table:NAS201}
			\centering
			\scalebox{0.85}{
				\begin{tabular}{l|c|cc|cc|cc|c}
					\hline
					\multicolumn{1}{c|}{\bf{Method}}& \bf{Search} & \multicolumn{2}{c|}{\bf{CIFAR-10}} & \multicolumn{2}{c|}{\bf{CIFAR-100}} & \multicolumn{2}{c|}{\bf{ImageNet-16-120}} & \bf{Search}\\
					&(seconds)& \it{validation} & \it{test} & \it{validation} & \it{test} & \it{validation} & \it{test}
					& \bf{Method}\\
					\hline
					RSPS \cite{li2019random} & $7587$ &$84.16\pm1.69$&$87.66\pm1.69$&$59.00\pm4.60$&$58.33\pm4.64$& $31.56\pm3.28$&$31.14\pm3.88$& random\\
					
					DARTS-V1 \cite{liu2018darts2} &$10890$ & $39.77\pm0.00$ & $54.30\pm0.00$ & $15.03\pm0.00$ &$15.61\pm0.00$ & $16.43\pm0.00$&$16.32\pm0.00$& gradient\\
					
					DARTS-V2 \cite{liu2018darts2} & $29902$ & $39.77\pm0.00$ & $54.30\pm0.00$ & $15.03\pm0.00$ &$15.61\pm0.00$& $16.43\pm0.00$&$16.32\pm0.00$& gradient\\
					
					GDAS \cite{dong2019searching} & $28926$ & $90.00\pm0.21$ & $93.51\pm0.13$ & $71.14\pm0.27$ &$70.61\pm0.26$& $41.70\pm1.26$&$41.84\pm0.90$& gradient\\
					
					SETN \cite{dong2019one} & $31010$ &$82.25\pm5.17$&$86.19\pm4.63$&$56.86\pm7.59$&$56.87\pm7.77$ &$32.54\pm3.63$&$31.90\pm4.07$& gradient\\
					
					ENAS \cite{pmlr-v80-pham18a} & $13314$ &$39.77\pm$0.00&$54.30\pm0.00$&$15.03\pm00$&$15.61\pm0.00$& $16.43\pm0.00$&$16.32\pm0.00$& RL\\
					
					EvNAS \cite{sinha2021evolving} & 22445 &88.98$\pm$1.40 & 92.18$\pm$1.11 & 66.35$\pm$2.59 & 66.74$\pm$3.08 & 39.61$\pm$0.72 & 39.00$\pm$0.44 & EA\\
					
					\bf{pEvoNAS} &\bf{4509}&\bf{90.54}$\pm$\bf{0.57}&\bf{93.63}$\pm$\bf{0.42}&\bf{69.28}$\pm$\bf{2.13}&\bf{69.05}$\pm$\bf{1.99}& \bf{40.00}$\pm$\bf{3.22}&\bf{39.98}$\pm$\bf{3.76}&\bf{EA}\\
					
					pEvoNAS (w/o inherit) &4509&86.86$\pm$2.50&89.83$\pm$3.16&67.90$\pm$2.09&68.21$\pm$2.48& 36.70$\pm$6.60&35.91$\pm$7.92&EA\\
					
					\hline
					ResNet & N/A &$90.83$&$93.97$&$70.42$&$70.86$&$44.53$&$43.63$& manual\\
					Optimal & N/A &$91.61$&$94.37$&$73.49$&$73.51$&$46.77$&$47.31$& N/A\\
					\hline
					
			\end{tabular}}
		\end{table*}
		
		\begin{figure*}[h]
			\centering
			\begin{center}
				\includegraphics[width=\linewidth]{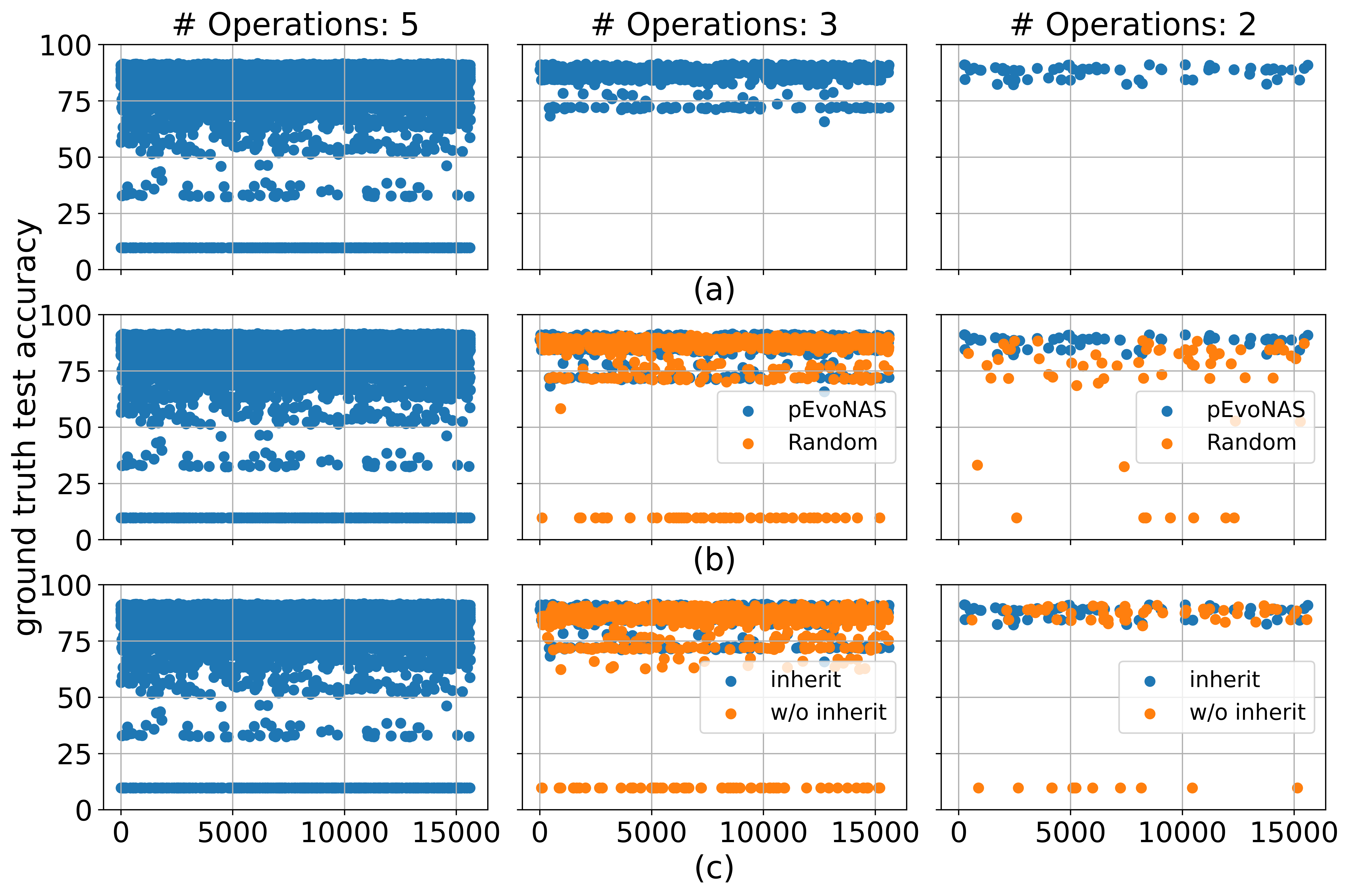}
			\end{center}
			\caption{Visualizing the search space by plotting the ground truth accuracies
				of all architectures in the search space.
				The \textit{x-axis} represents all 15,625 architectures in the
				search space S2 \cite{Dong2020NAS-Bench-201} and the \textit{y-axis}
				represents the true test accuracies.
				(a) Search space reduction using pEvoNAS.
				(b) Comparison of search space reduction using pEvoNAS and random search.
				(c) Comparison of search space reduction using pEvoNAS with weight
				inheritance (inherit) and no weight inheritance (w/o inherit).
			}
			\label{fig:ablation_search_space}
		\end{figure*}
		
		\subsection{Results}
		\label{results}
		\subsubsection{\textbf{Search Space 1 (S1):}}
		We performed 3 architecture searches on both CIFAR-10 and CIFAR-100 with
		different random number	seeds; their results are provided in
		Table~\ref{table:c10_s1} and Table~\ref{table:c100_s1}. The results show that
		the cells discovered by pEvoNAS on CIFAR-10 and CIFAR-100 achieve better results
		than those by human designed, RL based, gradient-based and EA-based methods.
		On comparing the computation time (or \textit{search cost}) measured in terms of
		\textit{GPU days}, we found that pEvoNAS performs the architecture search in
		significantly less time as compared to other EA-based methods while giving better
		search results. GPU days for any NAS method is calculated by multiplying the
		number of GPUs used in the NAS method by the execution time (reported in units
		of days). All the discovered architectures for S1 are provided in the
		supplementary.
		
		We followed	\cite{liu2018progressive}\cite{liu2018darts2}\cite{pmlr-v80-pham18a}
		\cite{real2019regularized}\cite{zoph2018learning} to
		compare the transfer capability of pEvoNAS with that of the other NAS methods,
		wherein the discovered architecture	on a dataset was transferred to another
		dataset (i.e. ImageNet) by retraining the architecture from scratch on the new
		dataset. The best discovered architectures from the architecture search on
		CIFAR-10 and CIFAR-100 (i.e. pEvoNAS-C10A and pEvoNAS-C100A) were then evaluated on
		the ImageNet dataset in mobile setting and the results are provided in
		Table~\ref{table:imagenet_s1}. The results show that the cells discovered by
		pEvoNAS on CIFAR-10 and CIFAR-100 can be successfully transferred to ImageNet,
		while using significantly less computational resources than EA based methods 
		
		\subsubsection{\textbf{Search Space 2 (S2):}}
		We performed 3 architecture searches each on 
		CIFAR-10, CIFAR-100 and	ImageNet-16-120 and their results are provided in
		Table~\ref{table:NAS201}. The results show that pEvoNAS outperforms most of the
		NAS methods	except GDAS \cite{dong2019searching} on CIFAR-100 and
		ImageNet-16-120. However, GDAS performs worse when the size of the search space
		increases as can be seen for S1 in Table~\ref{table:c10_s1}. In
		Figure~\ref{fig:s2_comparison}, we compare the progression of the search of
		pEvoNAS with that of other NAS methods. From the figure, we find that
		gradient-based method like DARTS, \cite{liu2018darts2}, suffers from overfitting
		problem wherein it converges to parameter-less operation, \textit{skip-connect}
		(i.e. a local optimum) \cite{chen2019progressive}\cite{Zela2020Understanding}
		\cite{Dong2020NAS-Bench-201}. In contrast, pEvoNAS does not get stuck to a
		local optimum architecture due to its stochastic nature. We also find that
		pEvoNAS converges to a solution much faster than other NAS methods.
		
		\section{Futher Analysis}
		\subsection{Visualizing the Architecture Search}
		\label{subsect:viz_arch_search}
		For analyzing the architecture search, we use S2 \cite{Dong2020NAS-Bench-201} to
		visualize the search process as it provides the true test accuracies of all the
		architectures in the search space. As illustrated in
		Figure~\ref{fig:ablation_search_space}(a), the search process is visualized by
		plotting all the architecures present in a given search space. From the figure, we
		find that the architecture search begins with the full search space (i.e. 5
		operations) and is then progressively reduced to smaller search spaces (i.e. 3
		operations and 2 Operations). More specifically, pEvoNAS reduces the search spaces
		to regions with high quality architectures and finally leading to the final search
		space (i.e. 2 operations) containing all architectures with top test accuracies. 
		
		\comm{
			\begin{table}[t]
				\centering
				\begin{tabular}{|c|c|c|c|}
					\multicolumn{4}{c}{\bf{Kendall Tau}}\\
					\hline
					$\#$ Operations& 5 & 3 & 2 \\
					\hline
					With weight inheritance & $0.16$&$0.17$&$0.35$\\		
					\hline
					Without weight inheritance & $0.16$&$0.15$&$0.22$\\		
					\hline
				\end{tabular}
				\caption{Correlation score (Kendall Tau) for supernet with weight inheritance
					and without weight inheritance in search spaces with reducing
					number of operations.
				}
				\label{table:wt_inherit}
			\end{table}
		}
		
		\subsection{Ablation Studies}
		\subsubsection{\textbf{Comparison with Random Search:}} Here, the search space, S1, is
		randomly reduced to the final search space (i.e. \# Operation: 2),
		$\Omega_{2}^{rand}$, and then a trained supernet is used to evaluated 100 random
		architectures in $\Omega_{2}^{rand}$. The best architecture is then returned as the
		final architecture, reported in Table~\ref{table:c10_s1} and Table~\ref{table:c100_s1}
		as pEvoNAS-C10rand and pEvoNAS-C100rand for CIFAR-10 and CIFAR-100 respectively.
		We found that the random search performs worse than pEvoNAS while taking lesser
		search time. To analyze the random search, we visualize the search space
		discovered using the random	search in S2, shown in
		Figure~\ref{fig:ablation_search_space}(b), by randomly reducing the search space to 
		smaller search spaces (i.e. 3 operations and 2 Operations). From the figure,
		we can see that the random search space reduction selects the search space with	both
		good and bad quality architectures which results into degraded output architecture.
		
		\subsubsection{\textbf{Effectiveness of Weight Inheritance:}}
		To illustrate the effectiveness of
		the weight inheritance of supernet, we perform the architecture search without weight
		inheritance in the search space S2 (given in Table~\ref{table:NAS201}) and found
		degraded search performance on all 3 datasets. We further analyze the differences
		in the search processes by plotting the search spaces discovered by pEvoNAS with
		inheritance and without inheritance respectively in
		Figure~\ref{fig:ablation_search_space}(c). From the figure, we see that the final
		search space discovered by not using weight inheritance contains both low and 
		high quality architectures as compared to only high quality architectures for
		pEvoNAS with weight inheritance. This reduction to lower quality search space shows
		the effectiveness of the weight inheritance of supernet during the search process.
		
		\section{Conclusion}
		\label{conclusion}
		The goal of this paper was to mitigate the noisy fitness estimation nature of the
		supernet by progressively reducing the search space to smaller regions of good
		quality architecture. This was achieved by using a trained supernet for architecture
		evaluation during the architecture search while using genetic algorithm to find regions
		in the search space with good quality architectures. The search then progressively
		reduces the search space to these regions and continues to search using a smaller
		supernet which inherits its weights from previous supernet. The use of trained
		supernet for evaluating the architectures in the population allowed us to skip
		the training of each individual architecture from sratch for its fitness
		evaluation and thus resulting in the reduced search time. We applied pEvoNAS to two
		different search spaces to show its effectiveness in generalizing to any
		cell-based search space. Experimentally, pEvoNAS reduced the search time of EA-based
		search methods significantly while achieving better results on CIFAR-10 and CIFAR-100
		datasets in S1 search space. We also visualized the search process using the NAS
		benchmark, NAS-Bench-201\cite{Dong2020NAS-Bench-201} and found that pEvoNAS
		progressively reduces the search space to smaller search spaces with top accuracy
		architectures.
	\comm{
	\subsection{Paper length}
	Papers, excluding the references section, must be no longer than eight pages in length.
	The references section will not be included in the page count, and there is no limit on the length of the references section.
	For example, a paper of eight pages with two pages of references would have a total length of 10 pages.
	{\bf There will be no extra page charges for \confName\ \confYear.}
	
	Overlength papers will simply not be reviewed.
	This includes papers where the margins and formatting are deemed to have been significantly altered from those laid down by this style guide.
	Note that this \LaTeX\ guide already sets figure captions and references in a smaller font.
	The reason such papers will not be reviewed is that there is no provision for supervised revisions of manuscripts.
	The reviewing process cannot determine the suitability of the paper for presentation in eight pages if it is reviewed in eleven.
	}
	{\small
		\bibliographystyle{ieee_fullname}
		\bibliography{references}
	}
	
\end{document}


\newcommand{\comm}[1]{}
	
	
	\title{Supplementary}
	
\maketitle

\comm{\begin{abstract}
Here, we present the settings used for the datasets in both the search spaces.
We also present the pseudocodes for the supernet training and the evolutionary
algorithm. Finally, we present all the discovered cells in S1 in
Figure~\ref{fig:searched_cells}.	
\end{abstract}
}

	\section{Dataset Settings}
	\label{sec:dataset_settings}
	The settings used for the datasets in \textbf{S1} are as follows:
	\begin{itemize}
		\item \textit{CIFAR-10}: We split 50K training images into two sets of size
		25K each, with one set acting as the training set and the other set as the
		validation set.
		\item \textit{CIFAR-100}: We split 50K training images into two sets. One set
		of size 40K images becomes the training set and the other set of size 10K
		images becomes the validation set.
	\end{itemize}
	The settings used for the datasets in \textbf{S2} are as follows:
	\begin{itemize}
		\item \textit{CIFAR-10}: The same settings as those used for S1 is used here
		as well.
		\item \textit{CIFAR-100}: The 50K training images remains as the training set
		and the 10K testing images are split into two sets of size 5K each, with one
		set acting as the validation set and the other set as the test set.
		\item \textit{ImageNet-16-120}: It has 151.7K training images, 3K validation
		images and 3K test images.
	\end{itemize}

	\section{Pseudocodes}
	The pseudocode for both training the supernet is given in
	Algorithm~\ref{algo:training_supernet}. It takes supernet, training data and total 
	number of epochs as inputs and outputs the trained supernet. The pseudocode for
	the evolutionary algorithm is given in Algorithm~\ref{algo:EA}.
	It uses the trained supernet to find the best architecture in the search space
	using validation data.

	\begin{algorithm}[t]
		\comm{https://arxiv.org/pdf/1903.03614.pdf}	
		\caption{TrainSupernet}	
		\label{algo:training_supernet}
		\SetAlgoLined
		\KwIn{Supernet $S^{\Omega_{op}}$, training data $\mathfrak{D}_{tr}$,
			total epochs $N_{epochs}$.
		}
		\KwOut{Trained supernet $S^{\Omega_{op}}$.}
		\For{ $\tau \gets 1$ to $N_{epochs}$ }{
			\For{each batch ($\mathfrak{B}$) in $\mathfrak{D}_{tr}$} {
				Update $\alpha$ to random values and send to $S^{\Omega_{op}}$\;
				Update weights of $S^{\Omega_{op}}$ using \textit{SGD}\;
			}
		}
	\end{algorithm}

	\begin{algorithm}[h]
		\comm{https://arxiv.org/pdf/1903.03614.pdf}	
		\caption{EA}	
		\label{algo:EA}
		\SetAlgoLined
		\KwIn{Search space $\Omega_{op}$, trained supernet $S^{\Omega_{op}}$,
			validation data	$\mathfrak{D}_{va}$, convergence number $N_{conv}$,
			population size	$N_{pop}$, mutation rate $r$, tournament size $T$.
		}
		\KwOut{Best architecture, $E_{best}$.}
		Initialize population, $P$ of size $N_{pop}$ for $\Omega_{op}$\;
			\While{ $E_{best}$ not repeated $N_{conv}$ times }{
				Evaluate each architecture in $P$ using $S^{\Omega_{op}}$ and	$\mathfrak{D}_{va}$\;
				Set $E_{best} \gets$ best architecture\;
				Copy $E_{best}$ to next generation population, $P_{next}$\;
				\For{ $i \gets 2$ to $N_{pop}$}{
					\tcc{Tournament Selection}
					Use tournament of size $T$ for selecting 2 parents\; 
					\tcc{Crossover}
					Use \textit{crossover} to create new child from the 2 parents for
					\textit{$P_{next}$}\;
					\tcc{Mutation}
					\For{each edge in \textit{child}}{
						\If{$uniformRandom(0,1) \leq r$} {
							Apply \textit{mutation} operation to the edge\;
						}
					}
				}
				$P \gets P_{next}$\;	
			}
	\end{algorithm}

	\section{Discovered Cells in S1}

	\begin{figure*}[t]
		\centering
		\subfloat[]{
			\includegraphics[width=0.45\linewidth]{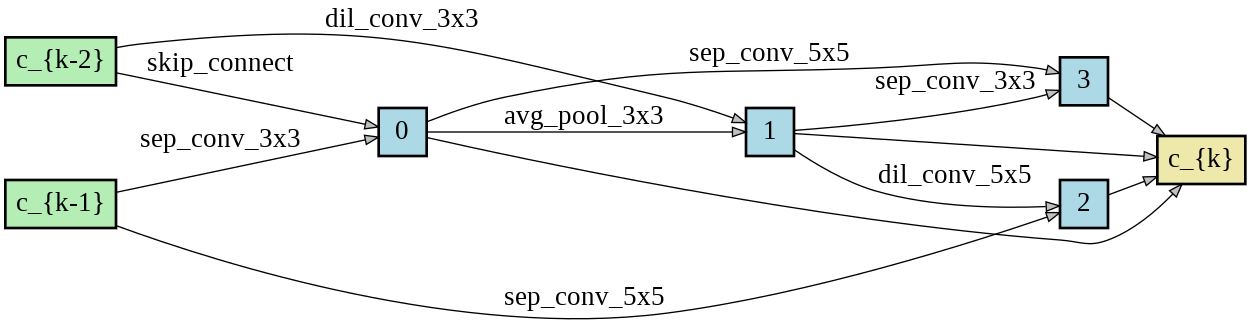}
		}
		\qquad
		\subfloat[]{
			\includegraphics[width=0.45\linewidth]{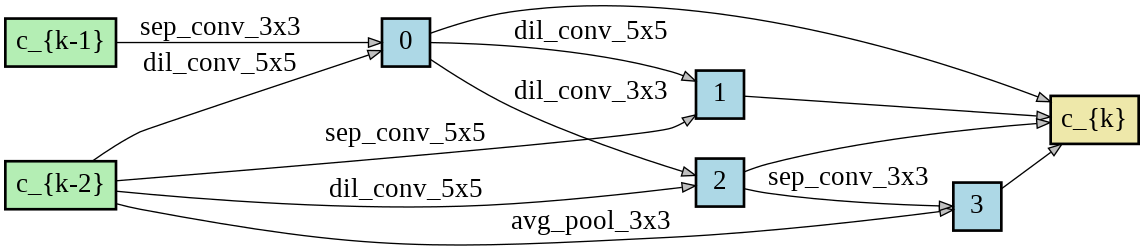}
		}
		\qquad
		\subfloat[]{
			\includegraphics[width=0.45\linewidth]{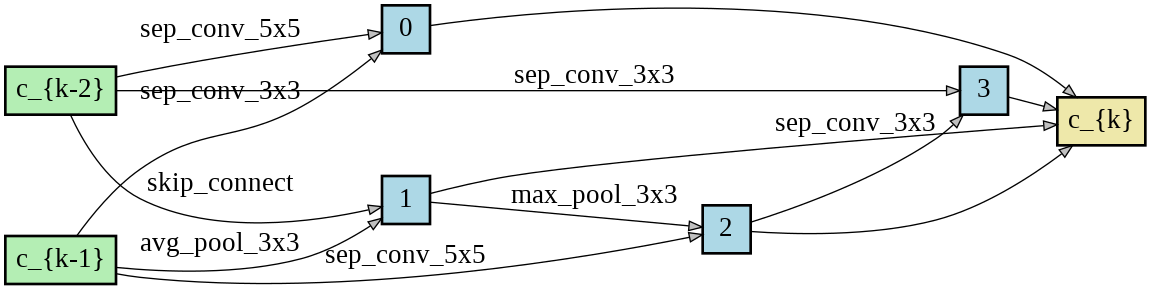}
		}
		\qquad
		\subfloat[]{
			\includegraphics[width=0.45\linewidth]{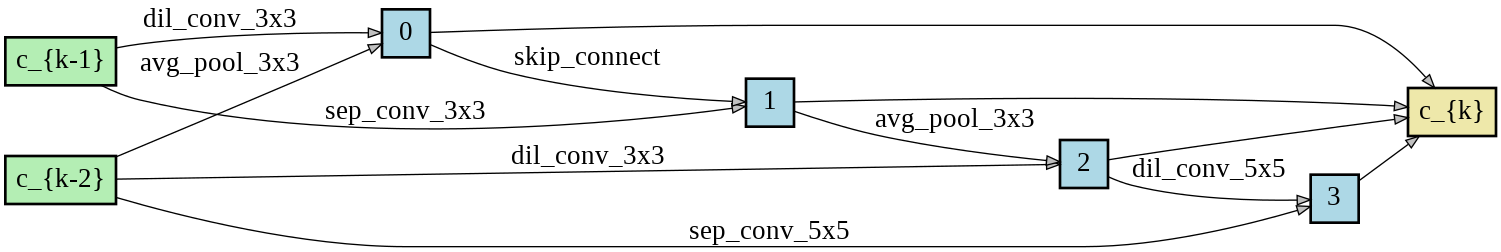}
		}
		\qquad
		\subfloat[]{
			\includegraphics[width=0.45\linewidth]{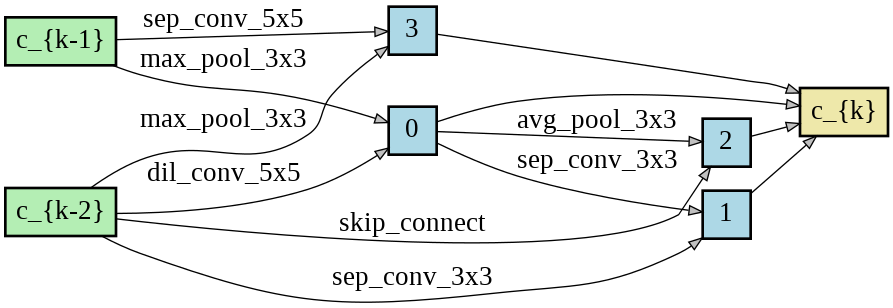}
		}
		\qquad
		\subfloat[]{
			\includegraphics[width=0.45\linewidth]{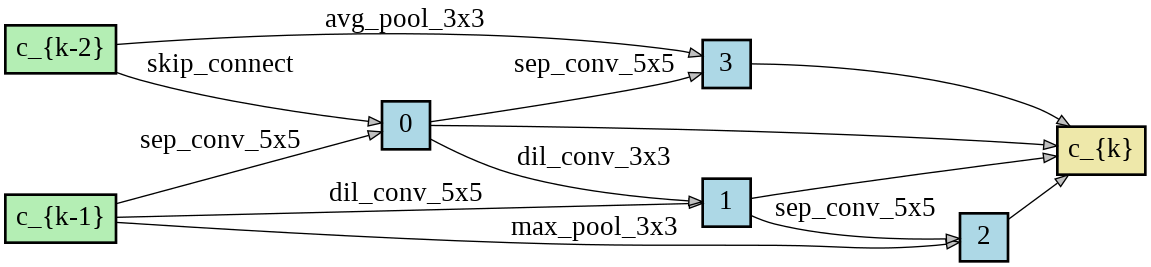}
		}
		\qquad
		\subfloat[]{
			\includegraphics[width=0.45\linewidth]{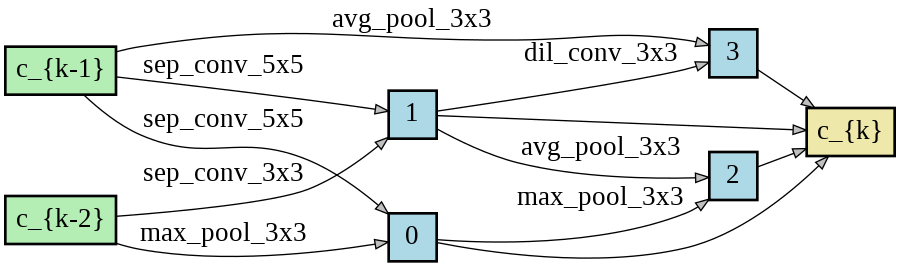}
		}
		\qquad
		\subfloat[]{
			\includegraphics[width=0.45\linewidth]{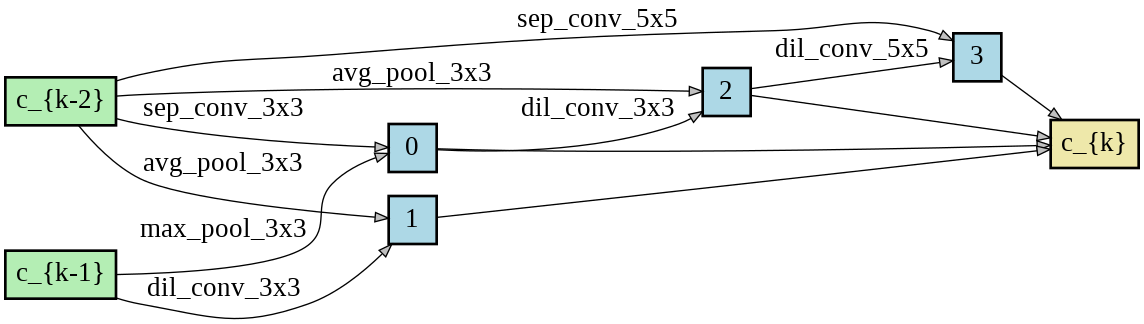}
		}	
		\qquad
		\subfloat[]{
			\includegraphics[width=0.45\linewidth]{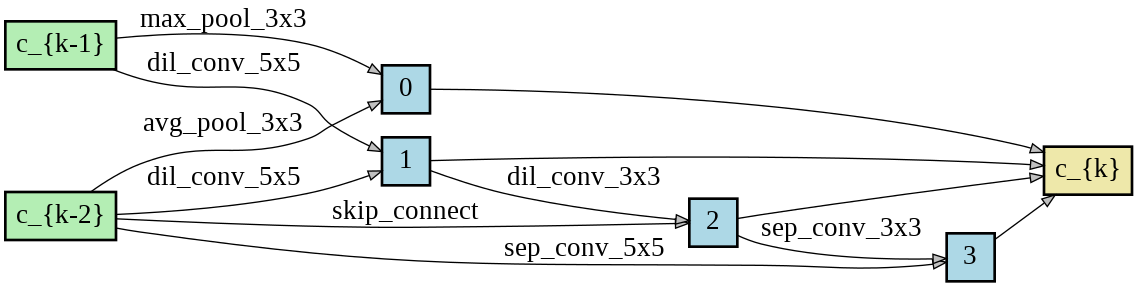}
		}
		\qquad
		\subfloat[]{
			\includegraphics[width=0.45\linewidth]{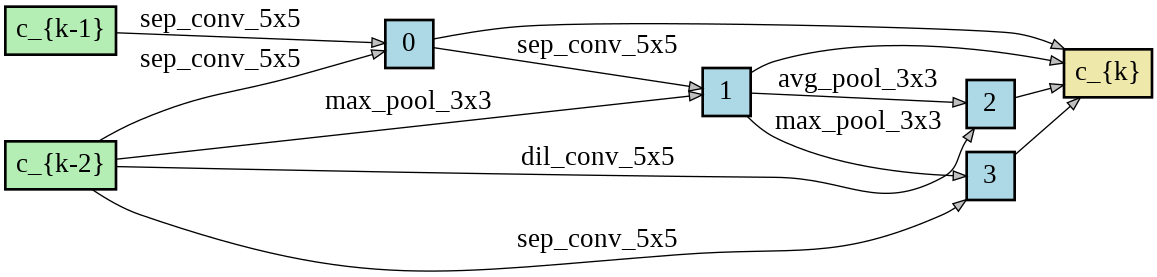}
		}
		\qquad
		\subfloat[]{
			\includegraphics[width=0.45\linewidth]{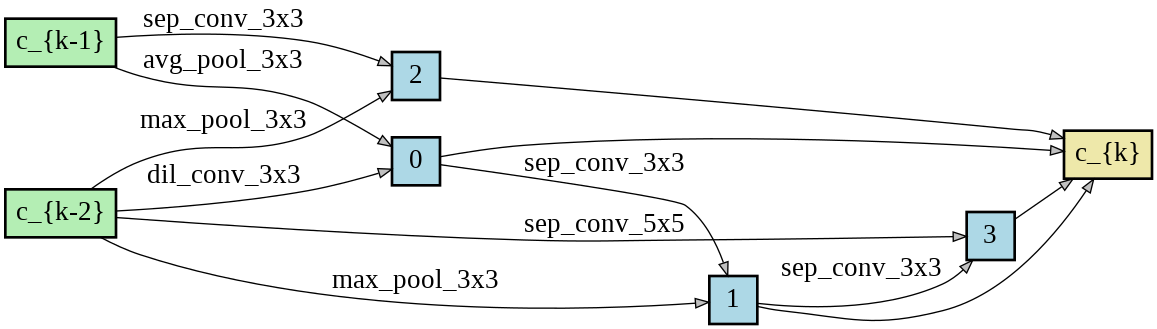}
		}
		\qquad
		\subfloat[]{
			\includegraphics[width=0.45\linewidth]{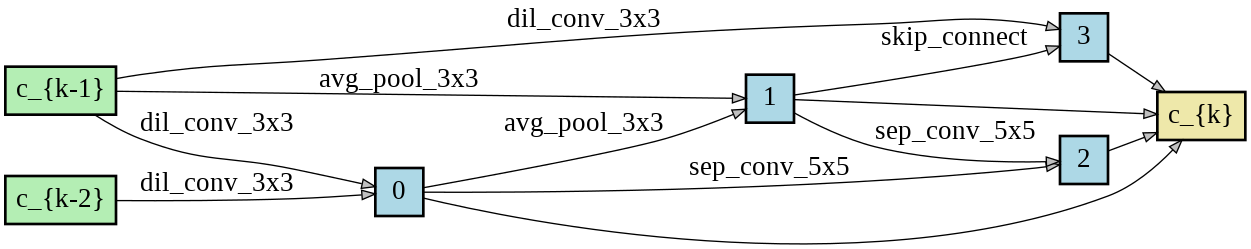}
		}
		
		\caption{Cells discovered by pEvoNAS-C10A (a) Normal cell (b) Reduction cell;
			by pEvoNAS-C10B (c) Normal cell (d) Reduction cell;
			by pEvoNAS-C10C (e) Normal cell (f) Reduction cell;
			by pEvoNAS-C100A (g) Normal cell (h) Reduction cell;
			by pEvoNAS-C100B (i) Normal cell (j) Reduction cell;
			by pEvoNAS-C100C (k) Normal cell (l) Reduction cell.}
		\label{fig:searched_cells}
	\end{figure*}
	